\documentclass[runningheads]{llncs}
\usepackage[T1]{fontenc}
\usepackage{graphicx}
\usepackage{amsmath,amssymb,amsfonts}
\usepackage{listings}
\usepackage{dsfont}
\usepackage{hyperref}

\usepackage{dsfont}
\usepackage{booktabs}
\usepackage{threeparttable}

\usepackage{xurl}

\begin{document}
\title{Stacking for Probabilistic Short-term Load Forecasting}
%
%
\author{Grzegorz Dudek \orcidID{0000-0002-2285-0327}}
\authorrunning{G. Dudek}
%
\institute{Electrical Engineering Faculty, Czestochowa University of Technology, Poland\\
\email{grzegorz.dudek@pcz.pl}}
\maketitle  
\begin{abstract}

In this study, we delve into the realm of meta-learning to combine point base forecasts for probabilistic short-term electricity demand forecasting. Our approach encompasses the utilization of quantile linear regression, quantile regression forest, and post-processing techniques involving residual simulation to generate quantile forecasts. Furthermore, we introduce both global and local variants of meta-learning. In the local-learning mode, the meta-model is trained using patterns most similar to the query pattern. 
Through extensive experimental studies across 35 forecasting scenarios and employing 16 base forecasting models, our findings underscored the superiority of quantile regression forest over its competitors.
\end{abstract}
\section{Introduction}

Ensembling stands out as a highly effective strategy for enhancing the predictive power of forecasting models. By combining predictions from multiple models, the ensemble approach consistently yields heightened accuracy. It leverages the strengths of individual models while mitigating their weaknesses, thereby extending predictive capacity by capturing a broader range of patterns and insights within the data. Additionally, ensembling accommodates the inclusion of multiple influential factors in the data generation process, alleviating concerns regarding model structure and parameter specification \cite{Wan22}. This comprehensive approach minimizes the risk associated with relying solely on a single model's limitations or biases, providing a more holistic representation of the data generation process.
Moreover, ensembling demonstrates remarkable robustness in handling outliers or extreme values and plays a crucial role in mitigating overfitting. It often achieves computational efficiency by leveraging parallel processing and optimization techniques.

Among the various methods for combining forecasts, the arithmetic average with equal weights emerges as a surprisingly robust and commonly used approach, often outperforming more complex weighting schemes \cite{Gen13}. Additionally, alternative strategies such as median, mode, trimmed means, and winsorized means have been explored \cite{Lic20}.

Linear regression serves as a valuable tool for assigning distinct weights to individual models, with weights estimated through ordinary least squares. These weights can effectively reflect the historical performance of base models \cite{Paw20}. They can be derived also from information criteria \cite{Kol11}, diversity of individual learners \cite{Kan22}, or time series characteristic features \cite{Mon20}.

While averaging and linear regression are prevalent methods for combining forecasts, they may fall short in capturing nonlinear and complex relationships between base models' forecasts and the target value. In such cases, machine learning (ML) models offer an alternative through stacking procedures \cite{Bab16}, \cite{Gas21}, which optimize forecast accuracy by learning the optimal combination of constituent forecasts in a data-driven manner.
The literature underscores the advantages of stacking generalization, including forecasting tasks involving time series with intricate seasonality, such as short-term load forecasting (STLF). For instance, \cite{Rib19} proposed utilizing wavelet neural networks (NNs) both as base forecasters and meta-learners for handling series with multiple seasonal cycles. In \cite{Div18}, the authors employed regression trees, random forests (RFs), and NNs as base models for STLF, with gradient boosting as a meta-learner. For a similar forecasting challenge, \cite{Suj22} combined RFs, long short-term memory (LSTM), deep NNs, and evolutionary trees with gradient boosting models as meta-learners, achieving significant forecast error reduction. In \cite{Gas21}, the authors compared forecast combination strategies, demonstrating the superior performance of stacking over methods like simple averaging and linear combination, showcasing its efficacy across diverse time series characteristics.

In this study, we propose a stacking approach to produce probabilistic STLF. 
STLF refers to the prediction of electricity demand over a relatively short period, typically ranging from a few hours to a few days ahead. Accurate STLF is crucial for efficient operation and planning of power systems, as it helps utility companies optimize generation, transmission, and distribution of electricity, leading to cost savings, improved reliability, and better utilization of resources.
Probabilistic forecasting in power systems is essential for addressing uncertainties stemming from factors like weather variations, load fluctuations, and unexpected events \cite{Hau19}. Power system operators rely on probabilistic forecasts to inform decisions, improve operational efficiency, manage risks, and ensure a dependable power supply. Likewise, in energy trading, probabilistic forecasting assists traders in understanding potential outcomes and uncertainties related to electricity prices, load demand, and generation availability \cite{Bey22}. 

Several instances of stacking for probabilistic STLF exist in the literature. For instance, \cite{He22a} employs a quantile regression LSTM meta-learner to combine point forecasts from tree-based models, including RF, gradient boosting decision tree, and light gradient boosting machine. It determines the probability density function (PDF) using kernel density estimation modified by Gaussian approximation of quantiles. In another study, \cite{Li19}, a Gaussian mixture distribution is utilized to combine probability density forecasts generated by various quantile regression models, such as Gaussian process regression, quantile regression NNs, and quantile regression gradient boosting. The results demonstrate the effectiveness of the proposed method in enhancing forecasting performance compared to methods employing simple averaging.

This study represents an extension of our previous conference paper \cite{Dud23}, which primarily focused on deterministic STLF. Here, our primary objective is to introduce methodologies for generating probabilistic STLF, building upon the point base forecasts.

Our study makes the following contributions:

\begin{enumerate}
\item 
Exploration of three approaches for combining point forecasts to produce probabilistic forecasts: post-processing techniques involving residual simulation, quantile linear regression, and quantile regression forest.
\item 
Investigation of global and local meta-learning strategies, which comprehensively capture both overarching data patterns and subtle nuances crucial for addressing complex seasonality.
\item 
Validation through extensive experimentation across 35 STLF problems marked by triple seasonality, using 16 distinct base models.
\end{enumerate}

The subsequent sections of this work are structured as follows. Section 2 defines the forecasting problem and discusses global and local meta-learning strategies. In Section 3, we provide a detailed description of the proposed meta-learners for probabilistic STLF. Section 4 presents application examples along with a thorough analysis of the achieved results. Finally, Section 5 outlines our key conclusions, summarizing the findings from this study.

\section{Problem statement}

The challenge of forecast combination involves the goal of finding a regression function, denoted as $f$ (meta-model). This function aggregates forecasts for time $t$ generated by $n$ forecasting models (base models) to create either point or probabilistic forecasts. 
A point forecast is a single-value prediction of a future outcome, represented by a single number. In contrast, probabilistic forecasts can take the form of predictive intervals (PI), a set of quantiles, or PDF, and this PDF can have a parametric or non-parametric form. In this work, we focus on generating probabilistic forecasts in the form of a set of quantiles.
For a continuous cumulative distribution function (CDF), $F(y|X=x)=P(Y \leq y|X=x)$, the $\alpha$-quantile is defined as:

\begin{equation}
Q_\alpha(x)=\inf\{y:F(y|X=x) \geq \alpha\}
\label{eqq}
\end{equation}
where $\alpha \in [0,1]$ denotes a nominal probability level.

Function $f$ can make use of all available information up to time $t-h$, where $h$ signifies the forecast horizon. However, in this study, we limit this information to the base forecasts, represented by vector $\hat{\textbf{y}}_t = [\hat{y}_{1,t}, ..., \hat{y}_{n,t}]$.  
 The combined quantile forecast is given as $\tilde{\textbf{q}}_t = f(\hat{\textbf{y}}_t; \boldsymbol{\theta}_t)$, where $\tilde{\text{\textbf{q}}}_t = [\tilde{q}_t(\alpha)]_{\alpha \in \Pi}$, $\Pi$ represents the assumed set of probabilities $\alpha$, and $\boldsymbol{\theta}_t$ denotes model parameters.

The class of regression functions $f$ encompasses a wide range of mappings, including both linear and nonlinear ones. The meta-model parameters can either remain static or vary over time. To enhance the performance of the meta-model, we employ an approach where the parameters are learned individually for each forecasting task, using a specific training set tailored for that task, represented as $\Phi = \{(\hat{\textbf{y}}_\tau, y_\tau)\}_{\tau \in \Xi}$. Here, $y_\tau$ denotes the target value, and $\Xi$ is a set of selected time indices from the interval $T = 1, ..., t-h$. 

The base forecasting models generate forecasts for successive time points $T=1, ..., t$. To obtain an ensemble forecast for time $t$, a meta-model can be trained using all available historical data from period $\Xi={\{1, ..., t-h\}}$, referred to as the global approach. This method allows the model to leverage all past information to generate a forecast for time point $t$.

In the alternative local mode, the goal is to train the meta-model locally around query pattern $\hat{\textbf{y}}_t$. To achieve this, $k$ most similar input vectors to $\hat{\textbf{y}}_t$ are selected and included into the local training set. The Euclidean metric is used to determine the nearest neighbors. The rationale behind adopting local training mode is rooted in the assumption that by focusing on a more narrowly defined segment of the target function --- as opposed to the broader scope of the global function -- there will be an enhancement in forecasting accuracy for the given query pattern. However, a crucial aspect of this approach is the determination of the optimal size for the local area, i.e. the choice of the number of neighbors, $k$.

\section{Meta-models for probabilistic STLF}

Quantile regression is concerned with estimating the conditional quantiles of a response variable. Unlike traditional regression that models the conditional mean of the response, quantile regression provides a more comprehensive view by estimating various quantiles.
This approach facilitates a thorough statistical analysis of the stochastic relationships among random variables, capturing a broader range of information about the distribution of the response variable.

In our research, we present three approaches for probabilistic STLF, each centered around the concept of quantile modeling.

\subsection{Quantile estimation through residual simulation (QRS)}

This method operates under the assumption that the distribution of residuals observed for historical data remains consistent with the distribution for future data. The process involves the following steps. Firstly, a meta-model for deterministic forecasting is trained, as outlined in \cite{Dud23}.
Then residuals are computed for historical data, which could encompass all training data, selected training patterns, or validation patterns. In our approach, we compute residuals for all training patterns, but note that in the local training mode, training patterns are chosen based on their similarity to the query pattern.

In the subsequent step, these residuals are added to the point forecasts, and a distribution function is fitted to the resulting values. To accomplish this, we employ a nonparametric kernel method, known for its flexibility compared to parametric alternatives. Once the distribution is estimated, quantiles are calculated using the inverse CDF.

Drawing from the findings reported in \cite{Dud23}, the RF model was chosen for deterministic forecasting. The formulation of this model is outlined as follows:

\begin{equation}
 f(\hat{\textbf{y}}) = \sum_{\tau \in \Xi}w_{\tau}(\hat{\textbf{y}})y_{\tau}
\label{eqdt}
\end{equation}
\begin{equation}
 w_{\tau}(\hat{\textbf{y}})=\frac{1}{p}\sum_{j=1}^p
 \frac{\mathds{1}{\{\hat{\textbf{y}}_{\tau} \in \ell_j(\hat{\textbf{y}})\}}}
 {\sum_{\kappa \in \Xi}\mathds{1}{\{\hat{\textbf{y}}_{\kappa} \in \ell_{j}(\hat{\textbf{y}})\}}} 
 \label{eqdt1}
\end{equation}
where $p$ is the number of trees in the forest, 
$\ell_{j}$ denotes the leaf that is obtained when dropping $\hat{\textbf{y}}$ down the $j$-th tree, and $\mathds{1}$ denotes the indicator function.

The RF response is the average of all training patterns $\hat{\textbf{y}}_{\tau}$ that reached the same leaves (across all trees) as query pattern $\hat{\textbf{y}}$. It approximates the conditional mean $E(Y|X = \hat{\textbf{y}})$ by a weighted mean over the observations of the response variable $Y$.

\subsection{Quantile linear regression (QLR)}

Classical linear regression, focused on minimizing sums of squared residuals, is adept at estimating models for conditional means. In contrast, quantile regression provides a powerful framework for estimating models for the entire spectrum of conditional quantile functions.

As observed by Koenker in \cite{Koe01}, quantiles can be conceptualized as the solution to a straightforward optimization problem using the pinball loss function:

\begin{equation}
L_\alpha(y, q) =
\begin{cases}
(y-q)\alpha & \text{if } y \geq q\\
(y-q)(\alpha-1)  &\text{if } y < q 
\end{cases}
\label{eqrho}
\end{equation}
where $y$ represents the true value, and $q$ is its predicted $\alpha$-quantile.

Thus, the linear model in the form of

\begin{equation}
 f(\hat{\textbf{y}}) = \sum_{i=1}^n{a_i\hat{y}_i} + a_0
 \label{eq1}
\end{equation}
where $a_0, ..., a_n$ are coefficients,
can effectively generate quantiles when the loss function is \eqref{eqrho}. The optimization problem is linear and can be efficiently solved using the interior point (Frisch-Newton) algorithm.

\subsection{Quantile regression forest (QRF)}

QRF extends the principles of RF to accommodate quantile regression \cite{Mei06}. Similar to traditional RF, QRF builds an ensemble of regression trees, with each tree constructed using a bootstrap sample of the training data. Notably, at each split, a random subset of predictor variables is considered, introducing an element of randomness and promoting diversity among the trees.

In contrast to standard RF, which estimates the conditional mean, QRF focuses on estimating the full conditional distribution, $E(\mathds{1}{\{Y \leq y\}}|X = \hat{\textbf{y}})$.
This distribution is approximated in QRF through the weighted mean across observations of 
$\mathds{1}{\{Y \leq y\}}$ \cite{Mei06}:

\begin{equation}
 \hat{F}(y|X = \hat{\textbf{y}})=
\sum_{\tau \in \Xi}w_{\tau}(\hat{\textbf{y}})\mathds{1}{\{y_{\tau} \leq y\}}
\label{eqdt2}
\end{equation}

To obtain estimates of the conditional quantiles $Q_\alpha(\hat{\textbf{y}})$, the empirical CDF determined by QRF in \eqref{eqdt2} is incorporated into \eqref{eqq}.

It is worth noting that the weights in \eqref{eqdt2} align with those used in RF, as defined in \eqref{eqdt1}.
The construction process of QRF mirrors that of RF, and QRF shares the same set of hyperparameters for tuning. The primary distinction between RF and QRF lies in how they treat each leaf: while RF retains only the mean of the observations falling into a leaf, QRF retains the values of all observations in the leaf, not just their mean. Subsequently, QRF leverages this information to assess the conditional distribution.


\section{Experimental Study}

In this section, we assess the effectiveness of our proposed stacking approaches for combining forecasts in the context of probabilistic STLF for 35 European countries. The time series under consideration exhibit triple seasonality, encompassing daily, weekly, and yearly patterns. The ensemble of base models comprises 16 forecasting models of diverse types, elaborated upon in Subsection 4.5.

\subsection{Dataset}

We gathered real-world data from the ENTSO-E repository (\url{www.entsoe.eu/data/power-stats}) for our study. The dataset comprises hourly electricity loads recorded from 2006 to 2018, spanning 35 European countries. This dataset provides a rich array of time series, each showcasing distinctive features such as levels and trends, variance, patterns of seasonal fluctuations across different periods (annual, weekly, and daily), and random fluctuations.

\subsection{Training and Evaluation Setup}

The forecasting base models underwent optimization and training on data spanning from 2006 to 2017. These models were then employed to generate hourly forecasts for the entire year of 2018, on a daily basis (for a detailed methodology, refer to \cite{Smy23}). To assess the effectiveness of the meta-models, we selected 100 specific hours (test hours) for each country from the latter half of 2018, evenly distributed across this period. The forecasts for these chosen hours were aggregated by the meta-models. 

Each meta-model was trained for every test hour to generate a vector of quantiles of probabilities $\alpha \in \Pi={0.01, 0.02, ..., 0.99}$.
The training utilized data from January 1, 2018, up to the hour just before the forecasted hour ($h=1$). The meta-models were trained in both global and local modes. The local modes involved training on the $k$ nearest training patterns to the query pattern, with $k \in K=\{20, 40, ..., 200, 250, 300\}$.

\subsection{Hyperparameters}

We conducted a comprehensive evaluation of the meta-models, considering different hyperparameter values:

\begin{itemize}
\item QRS: RF was utilized for generating point forecasts.
 After conducting preliminary simulations, we opt for default values for RF hyperparameters: number of predictors to select at random for each decision split $r=n/3$ (as recommended by the RF inventors), minimum number of observations per tree leaf $q=1$ (leading to overtrained trees, mitigated by combining them in the forest), and number of trees in the forest $p=100$.
Additionally, for estimating the forecast distribution, a nonparametric kernel method leveraging a normal kernel with the bandwidth optimized for normal densities was employed.
\item QLR: Implemented using Matlab code provided by Roger Koenker, solving a linear program via the interior point method (\url{www.econ.uiuc.edu/~roger/research/rq/rq}).
\item QRF: The minimum number of observations per leaf, $q$, was searched over the set $\{1, 5, 10, 15, 20, 30, 40, 50, 60\}$, while other hyperparameters remained the same as for RF used in QRS.
\end{itemize}

The proposed meta-models were implemented in Matlab 2022b, and experiments were conducted on a Microsoft Windows 10 Pro operating system, with an Intel(R) Core(TM) i7-6950x CPU @3.0 GHz processor, and 48 GB RAM.

\subsection{Evaluation metrics}

To assess the performance and effectiveness of the base models, we use the following metrics for point STLF:

\begin{itemize}
\item MAPE: mean absolute percentage error,
\item MdAPE: median of the absolute percentage error (more robust to outlier forecasts than MAPE),
\item MSE: mean squared error (more sensitive to outlier forecasts than MAPE),
\item MPE: mean percentage error (a measure of the forecast bias),
\item StdPE: standard deviation of the percentage error (a measure of the forecast dispersion).
\end{itemize}

For probabilistic STLF we apply:
\begin{itemize}
\item MPQRE: mean percentage quantile regression error, which is defined based on pinball loss function \eqref{eqrho} as follows: 

\begin{equation}
MPQRE=\frac{100}{N|\Pi|} \sum_{i=1}^N \sum_{\alpha \in \Pi} 
\frac{L_\alpha(y_i, \tilde{q}_i(\alpha))}{y_i}
\label{eqerr1}
\end{equation}
where $N$ is the number of forecasts.

This metric is expressed in percentage of true value $y_i$ to facilitate comparison across countries.

To provide a more comprehensive view of the results, we introduce two additional metrics: MdPQRE (median of PQRE), which is robust to outlier errors, and StdPQRE (standard deviation of PQRE), which measures the dispersion of errors.

\item ReFr: relative frequency, also known as calibration or coverage, defined as follows \cite{Mar22}:

\begin{equation}
ReFr(\alpha)=\frac{1}{N} \sum_{i=1}^N \mathds{1}{\{y_i \leq \tilde{q}_i(\alpha)}\}
\label{eqrf}
\end{equation}

The desired value of $ReFr(\alpha)$ is nominal probability level $\alpha$. In other words, the predicted $\alpha$-quantiles should exceed the realized values in $100\alpha\%$ of cases, ensuring an ReFr of $\alpha$. To assess the average deviation of $ReFr(\alpha)$ from the desired $\alpha$ across all $\alpha \in \Pi$, we define the mean absolute ReFr error:

\begin{equation}
MARFE=\frac{1}{|\Pi|} \sum_{\alpha \in \Pi} |ReFr(\alpha)-\alpha|
\label{eqmrf}
\end{equation}

 We calculate also the median and standard deviation of ARFE: MdARFE and StdARFE, respectively.


\item MPWS: mean percentage Winkler score defined as:

\begin{equation}
MPWS=\frac{100}{N} \sum_{i=1}^N 
\frac{WS(y_i, \tilde{q}_{l,i}, \tilde{q}_{u,i})}{y_i}
\label{eqerr2a}
\end{equation}

\begin{equation}
WS(y, \tilde{q}_l, \tilde{q}_u) =
\begin{cases}
(\tilde{q}_u- \tilde{q}_l) + \frac{2}{\alpha}(\tilde{q}_l - y) & \text{if } y < \tilde{q}_l\\
(\tilde{q}_u- \tilde{q}_l) & \text{if } \tilde{q}_l \leq y \leq \tilde{q}_u\\
(\tilde{q}_u- \tilde{q}_l) + \frac{2}{\alpha}(y-\tilde{q}_u) & \text{if } y > \tilde{q}_u\\
\end{cases}
\end{equation}

where $y$ represents the true value, $\tilde{q}_l$ and $\tilde{q}_u$ represent the predicted lower and upper quantiles defining the $100(1-\alpha)\%$ PI, $\alpha=\alpha_u-\alpha_l$, $\alpha_u$ and $\alpha_l$ define $\tilde{q}_l$ and $\tilde{q}_u$, respectively.

This measure evaluates PI, not a forecast distribution expressed by a set of quantiles as MPQRE and MARFE do. In this study, we assumed 90\% PI, $\alpha_l=0.05$ and $\alpha_u=0.95$. 
MPWS, like MPQRE, is a normalized measure to enable comparison results for different countries.

As in the case of MPQRE and MARFE, we define two additional metrics: MdPWS -- median of PWS, and StdPWS -- standard deviation of PWS.

\item inPI, belowPI, abovePI: percentage of observed values in PI, below PI and above PI.
This simple measure allows us to compare the results with the desired results: 90/5/5 in our case. 

\item QMAPE, QMdAPE: quality metrics specifically devised to evaluate point forecasts derived from predicted quantiles. These metrics operate under the assumption that the point forecast aligns with the 0.5-quantile (median). By contrasting these median-based metrics with conventional point forecasting metrics like MAPE and MdAPE, we can gauge the effectiveness of probabilistic forecasting meta-models in producing accurate point forecasts.

\end{itemize}

\subsection{Base forecasting models}

We utilized a diverse array of forecasting models as the base learners, encompassing statistical models, ML models, and various recurrent, deep, and hybrid NN architectures retrieved from \cite{Smy23}. This broad selection spans a spectrum of modeling techniques, each equipped with distinct mechanisms for capturing temporal patterns in data. The incorporation of this varied ensemble of models was intentional, aiming to ensure ample diversity among the base learners to enhance the overall forecasting performance. 

Table \ref{tabEr} presents the results of the base models averaged over 100 selected test hours across 35 countries. The models exhibit variations in MAPE from 1.70 to 3.83 and in MSE from 224,265 to 1,641,288. Among these, cES-adRNN emerges as the most accurate model based on MAPE, MdAPE, and MSE, while Prophet ranks as the least accurate.

\begin{table}[h]
\setlength{\tabcolsep}{13pt}
\caption{Point STLF quality metrics for the base models.}
\centering
\begin{tabular}{|l|r|r|r|r|r|}
\hline
 Base model & MAPE & MdAPE & MSE & MPE & StdPE \\
\hline
ARIMA & 2.86 & 1.82 & 777012 & 0.0556 & 4.60 \\
ES  & 2.83 & 1.79 & 710773 & 0.1639 & 4.64 \\
Prophet  & 3.83 & 2.53 & 1641288 & -0.5195 & 6.24 \\
N-WE & 2.12 & 1.34 & 357253 & 0.0048 & 3.47 \\
GRNN & 2.10 & 1.36 & 372446 & 0.0098 & 3.42 \\
MLP  & 2.55 & 1.66 & 488826 & 0.2390 & 3.93 \\
SVM  & 2.16 & 1.33 & 356393 & 0.0293 & 3.55 \\
LSTM & 2.37 & 1.54 & 477008 & 0.0385 & 3.68 \\
ANFIS & 3.08 & 1.65 & 801710 & -0.0575 & 5.59 \\
MTGNN & 2.54 & 1.71 & 434405 & 0.0952 & 3.87 \\
DeepAR  & 2.93 & 2.00 & 891663 & -0.3321 & 4.62 \\
WaveNet  & 2.47 & 1.69 & 523273 & -0.8804 & 3.77 \\
N-BEATS  & 2.14 & 1.34 & 430732 & -0.0060 & 3.57 \\
LGBM & 2.43 & 1.70 & 409062 & 0.0528 & 3.55 \\
XGB  & 2.32 & 1.61 & 376376 & 0.0529 & 3.37 \\
cES-adRNN & 1.70 & 1.10 & 224265 & -0.1860 & 2.57 \\
\hline
\end{tabular}
\begin{tablenotes}[flushleft]
      \tiny 
\item ARIMA: autoregressive integrated moving average model,
\item ES: exponential smoothing model,
\item Prophet: modular additive regression model with nonlinear trend and seasonal components,
\item N-WE: Nadaraya–Watson estimator,
\item GRNN: general regression NN,
\item MLP: perceptron with a single hidden layer and sigmoid nonlinearities,
\item SVM: linear epsilon-insensitive support vector machine,
\item LSTM: long short-term memory,
\item ANFIS: adaptive neuro-fuzzy inference system,
\item MTGNN: graph NN for multivariate time series forecasting,
\item DeepAR: autoregressive recurrent NN model for probabilistic forecasting,
\item WaveNet: autoregressive deep NN model combining causal filters with dilated convolutions,
\item N-BEATS: deep NN with hierarchical doubly residual topology,
\item LGBM: light gradient-boosting machine,
\item XGB: extreme gradient boosting algorithm,
\item cES-adRNN -- contextually enhanced hybrid and hierarchical model combining exponential smoothing and dilated recurrent NN with attention mechanism.
   
    \end{tablenotes}
\label{tabEr}
\end{table}

\subsection{Results}

\begin{figure}[h]%
\centering
\includegraphics[width=0.48\textwidth]{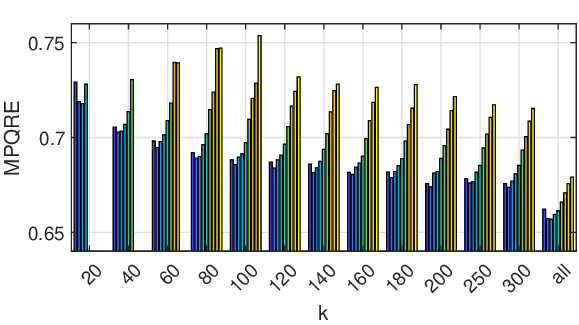}
\includegraphics[width=0.48\textwidth]{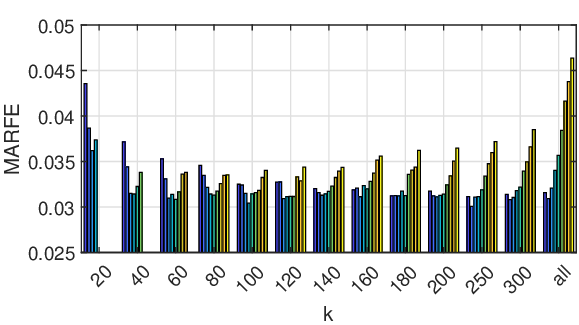}
\includegraphics[width=0.48\textwidth]{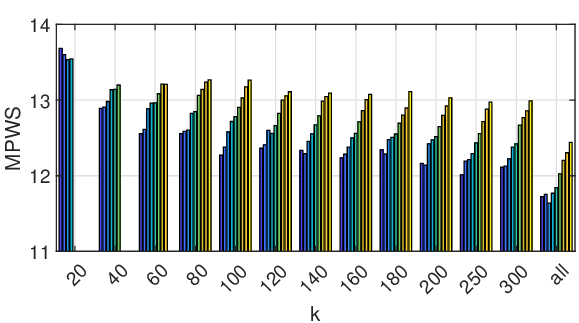}
\includegraphics[width=0.48\textwidth]{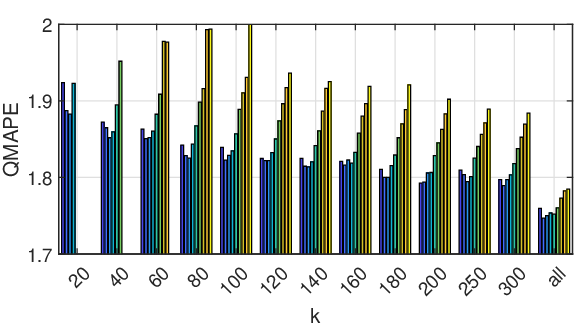}
\includegraphics[width=0.67\textwidth]{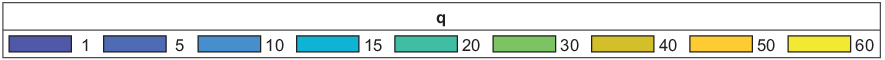}
\caption{Quality metrics for QRF.}\label{figp1}
\end{figure}

Fig. \ref{figp1} showcases the performance metrics of QRF across different training approaches and varying values of $q$ (the minimum number of observations per leaf). This visualization clearly illustrates that QRF attains its lowest error rates under global training conditions, especially when the $q$ parameter is set to 10 (resulting in the lowest values of MPQRE, MPWS, and QMAPE). However, it is important to note that MARFE draws slightly different conclusions. The MARFE metric for differnt $k$ exhibits less variability compared to other metrics, achieving its minimum value for $k=250$ and $q=5$ ($MARFE=0.0301$).

\begin{figure}[h]%
\centering
\includegraphics[width=0.48\textwidth]{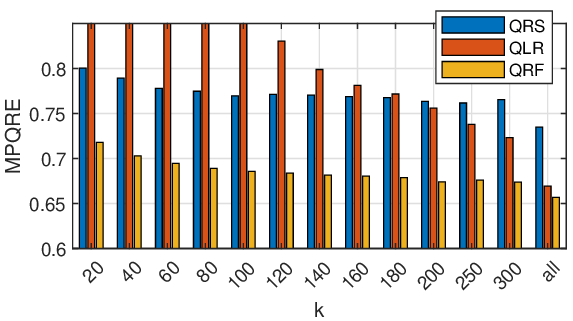}
\includegraphics[width=0.48\textwidth]{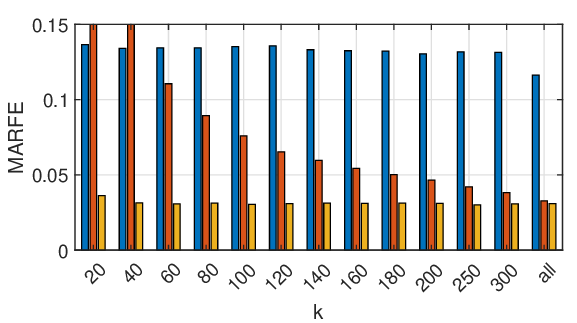}
\includegraphics[width=0.48\textwidth]{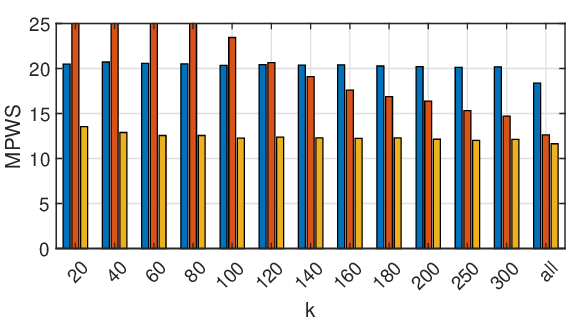}
\includegraphics[width=0.48\textwidth]{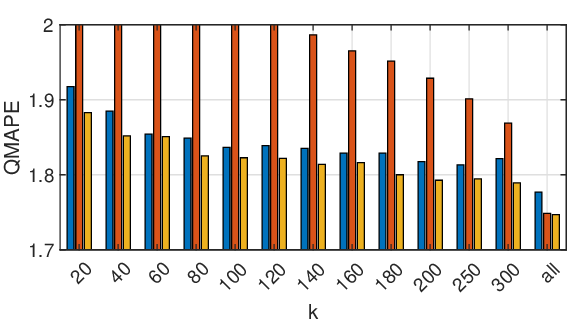}
\caption{Comparison of quality metrics for probabilistic STLF.}\label{figp2}
\end{figure}

Fig. \ref{figp2} provides a comparative evaluation of QRS, QRE, and QRF, focusing on their accuracy across global and local training methodologies. Notably, the QRE model displayed significant errors when trained on smaller datasets. As the size of the training dataset increased, all models demonstrated a gradual improvement in accuracy, with the most substantial enhancements observed in the global learning mode. It is worth highlighting that QRF stands out by achieving commendable results even with a limited training dataset size, as indicated by MARFE. Among the three models, QRF distinguishes itself by delivering superior accuracy and exhibiting minimal sensitivity to variations in the size of the training dataset. The subsequent results for QRF presented below are based on the global training mode with a $q$-value of 10.

Table \ref{tabPF} provides a concise summary of the quality metrics. 
Here, MARFE, MdARFE and StdARFE were determined based on $35 \cdot 99$ ReFr values calculated individually for each country.
A closer analysis of the metrics highlights QRF as the top-performing meta-model, exhibiting superior performance across most metrics.
Both QRF and QLR produce comparable PIs that closely resemble the ideal distribution, with a 90\% PI width and 5\% coverage on each side. In contrast, the PIs generated by QRS are notably narrower, providing coverage of less than 60\% on average, which falls short of the desired 90\% coverage.
Furthermore, when examining metrics like QMAPE and MdAPE, which are derived from the medians of the predicted distributions, it is evident that these values are larger compared to those generated by meta-models tailored for point forecasting (as observed in Table II in \cite{Dud23}).

\begin{table}[h]
\setlength{\tabcolsep}{12pt}
\centering
\caption{Probabilistic STLF quality metrics for meta-models.}
\begin{tabular}{|l|r|r|r|}
\hline
 Metric & QRS  & QLR  & QRF  \\
\hline
MPQRE   & 0.735 & 0.669 & \textbf{0.657} \\
MdPQRE  & \textbf{0.386} & 0.406 & 0.399 \\
StdPQRE & 1.066 & 0.913 & \textbf{0.877} \\
\hline
MARFE   & 0.1163 & 0.0327  & \textbf{0.0321}  \\
MdARFE  & 0.1100 & 0.0300 & \textbf{0.0200}  \\
StdARFE  & 0.0696 & \textbf{0.0268} &0.0295   \\
\hline
MPWS & 18.38 & 12.62 & \textbf{11.64} \\
MdPWS   & \textbf{4.09}  & 7.9  & 7.24 \\
StdPWS  & 38.44 & 23.69 & \textbf{20.99} \\
\hline
inPI & 59.26 & \textbf{90.34} & 90.77 \\
belowPI & 18.4 & \textbf{4.74}  & 4.23 \\
abovePI & 22.34 & 4.91 & \textbf{5.00} \\
\hline
QMAPE   & 1.78 & \textbf{1.75}  & \textbf{1.75}  \\
QMdAPE  & 1.12 & 1.12 & \textbf{1.09}  \\
\hline
\end{tabular}
\label{tabPF}
\end{table}

Table \ref{tabP2} provides a comprehensive breakdown of the MPQRE, MARFE and MPWS for each country, offering a detailed view of the model performance on a country-specific level. Notably, the QRF stands out as the most accurate model in terms of all metrics for the majority of countries included in the analysis. In contrast, QRS emerges as the most accurate model for just one country in this evaluation.

\begin{table}[!h]
\setlength{\tabcolsep}{6pt}
\caption{Probabilistic STLF quality metrics for each country.}
\label{tabP2}
\centering
\begin{tabular}{|@{}c|rrr|rrr|rrr@{}|}
\hline
   & \multicolumn{3}{c}{MPQRE} & \multicolumn{3}{|c|}{MARFE} &\multicolumn{3}{|c|}{MPWS} \\
Country   & QRS  & QRE  & QRF & QRS  & QRE  & QRF  & QRS   & QRE   & QRF   \\
\hline
AL  & 0.85 & \textbf{0.75} & 0.76 & 0.108 & 0.035 & \textbf{0.032} & 20.3 & 13.8 & \textbf{13.1} \\
AT  & 0.67 & \textbf{0.57} & 0.58 & 0.124 & \textbf{0.012}  & 0.030 & 16.7 & 10.9 & \textbf{9.2}  \\
BA  & 0.56 & 0.54  & \textbf{0.53} & 0.081 & \textbf{0.034}  & 0.044 & 13.7 & 10.0 & \textbf{9.8}  \\
BE  & 1.09 & 0.98  & \textbf{0.96} & 0.128 & 0.021 & \textbf{0.019} & 27.9 & 17.7 & \textbf{17.6} \\
BG  & 0.66 & \textbf{0.58} & \textbf{0.58} & 0.112 & 0.029 & \textbf{0.015} & 14.9 & \textbf{10.2}  & 10.3 \\
CH  & 1.23 & \textbf{1.06} & \textbf{1.06} & 0.149 & 0.032 & \textbf{0.018} & 30.4 & \textbf{19.1}  & 19.4 \\
CZ  & 0.54 & \textbf{0.51} & \textbf{0.51} & 0.125 & \textbf{0.021}  & 0.024 & 14.3 & 10.0 & \textbf{9.1}  \\
DE  & 0.51 & \textbf{0.44} & 0.46 & 0.108 & \textbf{0.028}  & 0.036 & 12.3 & 8.1  & \textbf{7.8}  \\
DK  & 0.86 & \textbf{0.73} & 0.75 & 0.135 & \textbf{0.040}  & 0.045 & 20.2 & \textbf{13.6}  & 13.7 \\
EE  & 0.81 & 0.72  & \textbf{0.70} & 0.091 & \textbf{0.048}  & 0.057 & 18.9 & \textbf{13.4}  & 13.4 \\
ES  & 0.45 & \textbf{0.37} & 0.39 & 0.127 & 0.026 & \textbf{0.020} & 10.2 & \textbf{6.6} & 6.8 \\
FI  & 0.51 & \textbf{0.44} & \textbf{0.44} & 0.113 & \textbf{0.016}  & 0.026 & 12.6 & 7.5  & \textbf{7.1}  \\
FR  & 0.57 & \textbf{0.52} & \textbf{0.52} & 0.088 & \textbf{0.019}  & 0.030 & 14.4 & 10.8 & \textbf{9.9}  \\
GB  & 1.40 & 1.32  & \textbf{1.26} & 0.106 & 0.022 & \textbf{0.016} & 35.9 & 22.3 & \textbf{21.8} \\
GR  & 0.84 & \textbf{0.71} & 0.77 & 0.111 & 0.047 & \textbf{0.037} & 22.5 & \textbf{13.5}  & 13.8 \\
HR  & 0.94 & \textbf{0.81} & \textbf{0.81} & 0.122 & 0.029 & \textbf{0.024} & 24.9 & \textbf{13.9}  & 14.5 \\
HU  & 0.74 & 0.65  & \textbf{0.62} & 0.146 & 0.029 & \textbf{0.027} & 20.4 & 13.5 & \textbf{10.8} \\
IE  & 0.52 & 0.56  & \textbf{0.51} & 0.116 & 0.022 & \textbf{0.020} & 12.1 & 10.9 & \textbf{8.7}  \\
IS  & 0.56 & \textbf{0.46} & \textbf{0.46} & 0.198 & \textbf{0.027}  & 0.059 & 15.0 & \textbf{10.0}  & 10.5 \\
IT  & 0.81 & 0.74  & \textbf{0.69} & 0.126 & 0.032 & \textbf{0.030} & 21.7 & 14.4 & \textbf{10.8} \\
LT  & 0.58 & 0.55  & \textbf{0.53} & 0.089 & 0.037 & \textbf{0.033} & 15.8 & 10.3 & \textbf{8.7}  \\
LU  & \textbf{0.67} & 0.90  & 0.69 & \textbf{0.037} & 0.068 & 0.066 & \textbf{12.1} & 20.6 & 14.7 \\
LV  & 0.62 & \textbf{0.57} & \textbf{0.57} & 0.126 & 0.048 & \textbf{0.026} & 13.9 & 9.2  & \textbf{8.2}  \\
ME  & 0.98 & 0.91  & \textbf{0.90} & 0.122 & 0.022 & \textbf{0.018} & 23.4 & 17.1 & \textbf{15.6} \\
MK  & 1.50 & \textbf{1.30} & \textbf{1.30} & 0.125 & \textbf{0.031}  & 0.033 & 40.0 & 23.7 & \textbf{21.1} \\
NL  & 0.65 & \textbf{0.55} & 0.59 & 0.108 & 0.050 & \textbf{0.046} & 16.0 & 11.3 & \textbf{11.0} \\
NO  & 0.73 & \textbf{0.62} & 0.65 & 0.125 & \textbf{0.029}  & 0.032 & 19.0 & \textbf{10.1}  & 10.4 \\
PL  & 0.58 & 0.51  & \textbf{0.50} & 0.119 & 0.035 & \textbf{0.024} & 15.8 & 11.0 & \textbf{9.1}  \\
PT  & 0.51 & \textbf{0.46} & 0.47 & 0.117 & \textbf{0.044}  & 0.049 & 11.8 & 9.0  & \textbf{8.7}  \\
RO  & 0.54 & \textbf{0.44} & 0.45 & 0.109 & 0.049 & \textbf{0.037} & 13.6 & 8.0  & \textbf{7.7}  \\
RS  & 0.53 & 0.51  & \textbf{0.50} & 0.106 & \textbf{0.032}  & 0.037 & 10.8 & 9.1  & \textbf{8.6}  \\
SE  & 0.70 & 0.67  & \textbf{0.63} & 0.112 & \textbf{0.036}  & 0.049 & 16.9 & 12.4 & \textbf{10.5} \\
SI  & 0.69 & 0.73  & \textbf{0.64} & 0.112 & 0.037 & \textbf{0.028} & 16.4 & 12.8 & \textbf{9.9}  \\
SK  & 0.59 & 0.54  & \textbf{0.52} & 0.131 & 0.034 & \textbf{0.012} & 15.9 & 11.3 & \textbf{9.8}  \\
TR  & 0.78 & 0.72  & \textbf{0.71} & 0.114 & 0.024 & \textbf{0.023} & 22.8 & 16.2 & \textbf{15.4} \\
\hline
\end{tabular}
\end{table}

\begin{figure}[h]
	\centering
	\includegraphics[width=0.28\textwidth]{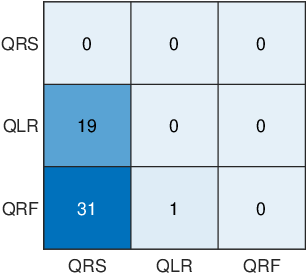}
  \caption{Results of the Diebold-Mariano tests for MPQRE.} 
	\label{fig4a}
\end{figure}

For a more robust assessment of model performance, a Diebold-Mariano test \cite{Die95} was executed to evaluate the statistical significance of differences between the probabilistic forecasts generated by each pair of models, considering individual country errors. The results based on MPQRE are visually depicted in Fig. \ref{fig4a}, where the diagram illustrates the instances in which the model represented on the y-axis is statistically more accurate than the model represented on the x-axis. It is worth highlighting that the statistical analysis reveals QRF's superior accuracy over QRS in a substantial majority, specifically 31 out of 35, of the considered countries. Additionally, QLR outperforms QRS in terms of accuracy in 19 countries. Interestingly, QRF surpasses QLR in accuracy for just one country in this comparison.

\begin{figure}[h]%
\centering
\includegraphics[width=0.32\textwidth]{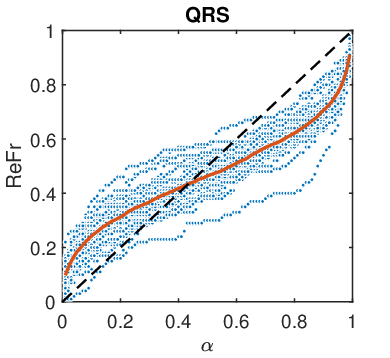}
\includegraphics[width=0.32\textwidth]{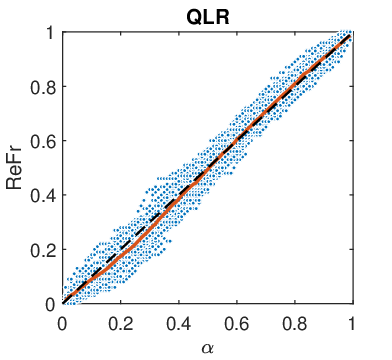}
\includegraphics[width=0.32\textwidth]{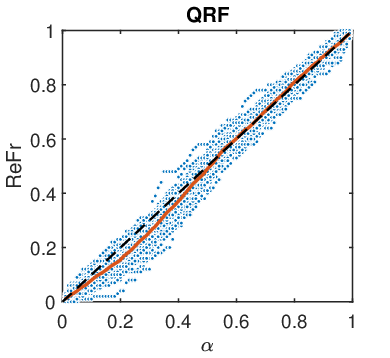}
\caption{Relative frequency.}\label{figp2a}
\end{figure}

The relative frequency (ReFr) charts illustrated in Fig. \ref{figp2a} provide a comprehensive view of the probabilistic forecasting outcomes. The desired value of $ReFr(\alpha)$ is nominal probability level $\alpha$, shown with the dashed line. ReFrs calculated for each country and each $\alpha \in \Pi$ are represented by dots, while the red line signifies the mean ReFr value across all countries.
Observing this figure, it is evident that QRS exhibits the most distorted ReFr distribution, indicating excessively narrow PIs.
In contrast, ReFrs for QLR and QRF are less dispersed and closely aligned with the desired values, with the average line almost coinciding with the dashed line.
It is worth noting that MARFE presented in Table \ref{tabPF} serves as an indicator of the average deviation of ReFr values for each $\alpha \in \Pi$ and each country (dots) from the desired value (dashed line).

\section{Conclusion}\label{sec13}

The practice of forecast combination has gained widespread recognition as an effective strategy for enhancing forecast accuracy and reliability. In this study, we focused on combining point forecasts using various stacking approaches to produce probabilistic forecasts. We introduced three approaches capable of generating quantile forecasts based on the point forecasts of base models. Notably, in the domain of short-term load forecasting, the quantile regression forest outperformed quantile linear regression slightly and demonstrated significant improvement over a method reliant on quantile estimation through residual simulation.

Our future research endeavors will be concentrated on the development of advanced machine learning models and approaches specifically tailored for forecast combination. This pursuit aims to further enhance predictive capabilities within bagging and boosting scenarios, ultimately advancing the field of forecasting.

%
%
%
\bibliographystyle{splncs04}
\bibliography{sn-bibliography}
%




\end{document}